\documentclass[letterpaper, 10 pt, conference]{ieeeconf}  

\IEEEoverridecommandlockouts                              

\usepackage{amsthm}
\usepackage{graphicx}
\usepackage{caption}
\usepackage{subcaption}
\usepackage[linesnumbered,ruled,vlined]{algorithm2e}
\usepackage{algorithmicx}
\usepackage{amsmath}
\theoremstyle{plain}
\usepackage{mathtools}
\usepackage{epsfig}
\usepackage{amssymb}
\usepackage{epstopdf}
\usepackage{multicol}
\usepackage{multirow}
\usepackage{array}
\usepackage{url}
\usepackage{color}
\usepackage{float}
\usepackage{wrapfig}

\title{\LARGE \bf
Deeply Informed Neural Sampling for Robot Motion Planning 
}

\author{Ahmed H. Qureshi and Michael C. Yip
\thanks{A. H. Qureshi and M. C. Yip are with Department of Electrical and Computer Engineering, University of California San Diego, La Jolla, CA 92093 USA. {\tt\small \{a1qureshi, yip\}@ucsd.edu}}%
}

\begin{document}

\maketitle
\thispagestyle{empty}
\pagestyle{empty}

\begin{abstract}
Sampling-based Motion Planners (SMPs) have become increasingly popular as they provide collision-free path solutions regardless of obstacle geometry in a given environment. However, their computational complexity increases significantly with the dimensionality of the motion planning problem. Adaptive sampling is one of the ways to speed up SMPs by sampling a particular region of a configuration space that is more likely to contain an optimal path solution. Although there are a wide variety of algorithms for adaptive sampling, they rely on hand-crafted heuristics; furthermore, their performance decreases significantly in high-dimensional spaces. In this paper, we present a neural network-based adaptive sampler for motion planning called Deep Sampling-based Motion Planner (DeepSMP). DeepSMP generates samples for SMPs and enhances their overall speed significantly while exhibiting efficient scalability to higher-dimensional problems. DeepSMP's neural architecture comprises of a Contractive AutoEncoder which encodes given workspaces directly from a raw point cloud data, and a Dropout-based stochastic deep feedforward neural network which takes the workspace encoding, start and goal configuration, and iteratively generates feasible samples for SMPs to compute end-to-end collision-free optimal paths. DeepSMP is not only consistently computationally efficient in all tested environments but has also shown remarkable generalization to completely unseen environments. We evaluate DeepSMP on multiple planning problems including planning of a point-mass robot, rigid-body, 6-link robotic manipulator in various 2D and 3D environments. The results show that on average our method is at least 7 times faster in point-mass and rigid-body case and about 28 times faster in 6-link robot case than the existing state-of-the-art.  
\end{abstract}
\section{Introduction}
Sampling-based Motion Planners (SMPs) have emerged as a promising framework for solving high-dimensional, constrained motion planning problems \cite{lavalle1998rapidly} \cite{karaman2011sampling}. SMPs ensure probabilistic completeness, which implies that a probability of finding a feasible path solution, if one exists, approaches to one as the limit of the number of randomly drawn samples from an obstacle-free space increases to infinity \cite{karaman2011sampling}. However, despite their ability to compute motion plans irrespective of the obstacles geometry, these methods exhibit slow convergence to computing path solutions due to their reliance on the extensive exploration of a given obstacle-free configuration space \cite{qureshi2016potential} \cite{qureshi2018motion}. Recent research shows that biasing a sample distribution towards the region with high probability of finding a path solution can considerably enhance the performance of classical single-query SMPs such as  RRT and RRT*  \cite{qureshi2016potential}. To the best of our knowledge, there does not exist any effective and reliable solution that uses the knowledge from the past planning problems to bias the sample distributions towards the region of the configuration space containing an optimal path solution.  

In this paper, we propose a neural network-based adaptive sampler that generates samples in particular regions of a configuration space where there is likely to exist an optimal path solution. Our method consists of two neural models, i.e., an obstacle-space encoder and random samples generator. We use a Contractive AutoEncoder (CAE) \cite{rifai2011contractive} for the encoding of an obstacle-space into an invariant, robust feature space. A samples generator that comprises a Dropout-based \cite{srivastava2014dropout} stochastic Deep Neural Network (DNN) that takes the obstacle-space encoding, start and goal configuration as an input, and generates samples distributing over the region of configuration space containing the path solutions. We evaluate our method on various complex motion planning tasks such as planning of a rigid-body (piano-mover problem) and 6 degree-of-freedom (DOF) robotic arm (UR6), and planning through narrow passages. We also benchmark our method against existing biased-sampling based state-of-the-art SMPs including Informed-RRT* \cite{gammell2014informed} and Batch Informed Trees (BIT*) \cite{gammell2015batch}. The results show that our algorithm generates samples that enable unbiased SMPs such as RRT* to compute near-optimal paths in a considerably lesser computational time than BIT* and Informed-RRT*. 

\section{Related Work} 
Many biased-sampling heuristics have been proposed to enhance the computational speed of RRT \cite{lavalle1998rapidly} and its variants. For instance, Rickert et al. \cite{rickert2008balancing} used gradient information to balance exploration and exploitation. Urmson and Simmons method \cite{urmson2003approaches} heuristically biased samples in RRT while Ferguson and Stentz \cite{ferguson2006anytime} presented the anytime RRT algorithm by using multiple independent RRTs. Although these methods are useful, they lack asymptotic optimality due to the underlying RRT algorithm. 

RRT* \cite{karaman2011sampling} extends RRTs to guarantee asymptotic optimality by incrementally rewiring the RRT graph connections such that the shortest path is asymptotically guaranteed \cite{karaman2011sampling}.  However, to determine an $\epsilon$-near optimal path in $d\in \mathbb{N}$ dimensions, roughly $O(1/\epsilon^d)$ samples are required, which makes RRT* no better than grid search methods \cite{hauser2015lazy}. Likewise, experiments in \cite{qureshi2016potential} \cite{gammell2014informed} also confirmed that RRT* exhibits slow convergence rates to optimal path solution in higher-dimensional spaces. The following sections discusses various existing biased/adaptive sampling methods to speed up the convergence rate of SMPs to compute optimal/near-optimal path solution.

\subsection{Adaptive Sampling Methods}
Gammell et al. \cite{gammell2014informed} proposed the Informed-RRT* algorithm which takes an initial solution from RRT* algorithm to define an ellipsoidal region from which new samples are drawn to minimize the initial solution for a given cost function. Although Informed-RRT* demonstrated enhanced convergence towards an optimal solution, this method suffers in situations where finding an initial path solution takes most of the computation time. To address this limitation, Gammell et al. proposed Batch Informed Trees (BIT*) \cite{gammell2015batch}. BIT* is an incremental graph search technique where an ellipsoidal subset, containing configurations to update the graph, is incrementally enlarged. BIT* is shown empirically to outperform prior methods such as RRT* and Informed-RRT*. However, confining a graph search to ellipsoidal region slows down the performance of an algorithm in maze-like scenarios especially where the start and goal configurations are very close to each other, but the path among them traverses a complicated maze stretching waypoints far away from the goal. Furthermore, such a method would not translate to non-stationary environments or unseen environments. 

\subsection{Learning-based Search Methods}
Many approaches exist that use learning to improve classical SMPs computationally. A recent method called a Lightning Framework \cite{berenson2012robot} stored paths into a lookup table and used a learned heuristic to write new paths as well as to read and repair old paths.  Another similar framework by Coleman et al. \cite{coleman2015experience} is an experience-based strategy to cache experiences in a graph instead of individual trajectories. Although these approaches exhibit superior performance in higher-dimensional spaces when compared to conventional planning methods, lookup tables are memory inefficient and incapable of generalizing well to new planning problems.  Zucker et al. \cite{zucker2008adaptive} proposed a reinforcement learning-based method to bias samples in discretized workspaces. However, reinforcement learning-based approaches are known for their slow convergence as they require a large number of interactive experiences. 

\begin{figure*}
    \centering
   \begin{subfigure}[b]{0.28\textwidth}
       \includegraphics[height=5.5cm]{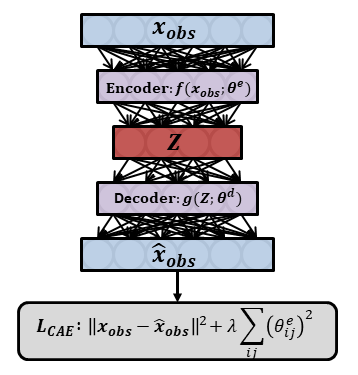}
       \caption{Offline: CAE}
    \end{subfigure}
     \begin{subfigure}[b]{0.25\textwidth}
        \includegraphics[height=5.5cm]{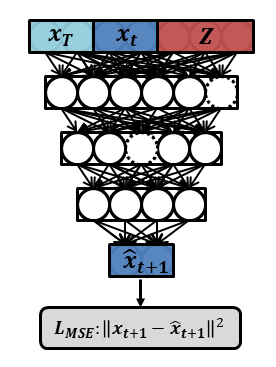}
        \caption{Offline: DeepSampler}
    \end{subfigure}
    \begin{subfigure}[b]{0.25\textwidth}
        \includegraphics[height=5.5cm]{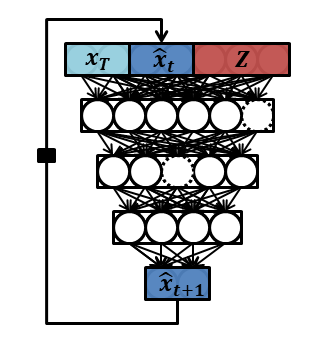}
        \caption{Online: Neural Sampling}
    \end{subfigure}
    \caption{DeepSMP consists of two neural models, a Contractive AutoEncoder (CAE), and a stochastic deep feedforward neural network (DeepSampler). These models are trained offline and are used to generate samples during online execution incrementally.}\label{DeepSMP}
\end{figure*}
\section{Problem Definition}
This section presents the notations we will be using in this paper, along with the definitions of fundamental motion planning problems addressed by our work.

Let $S$ be a list of finite length $N \in \mathbb{N}$ then $S_i$ is a mapping from a given index $i\in \mathbb{N}$ to an element of $S$ at $i$-th index. For algorithms described in our paper, $S_0$ and $S_T$  corresponds to the initial and last elements of a list, respectively. Let a given state space be denoted as $X \subset \mathbb{R}^d$, where $d \in \mathbb{N}_{\geq2}$ denotes the dimension of a state space. The collision and collision-free state spaces are denoted as $X_\mathrm{obs} \subset X$ and $X_\mathrm{free}= X \backslash X_\mathrm{obs}$, respectively. Let the initial state and goal region be represented as  $x_\mathrm{init} \in X_\mathrm{free}$ and $X_\mathrm{goal} \subset X_\mathrm{free}$, respectively. Let a trajectory be denoted as a non-empty finite-length list $\sigma:[0,T] \subset X$. For a given path planning problem, a trajectory $\sigma$ is said to be \textit{feasible} if it connects $x_\mathrm{init}$ and $ x \in X_\mathrm{goal}$, i.e. $\sigma_0=x_\mathrm{init}$ and $\sigma_T \in X_\mathrm{goal}$, and a path formed by connecting all consecutive states in $\sigma$ lies entirely in the obstacle-free space $X_\mathrm{free}$ i.e.,\\ \\
\textbf{Problem 1 (Feasible Path Planning)}\textit{ Given a triplet $\{X,X_\mathrm{free},X_\mathrm{obs}\}$, an initial state $x_\mathrm{init}$ and a goal region $X_\mathrm{goal}\subset X_\mathrm{free}$, find a path $\sigma:[0,T] \rightarrow X_\mathrm{free}$ such that $\sigma_0=x_\mathrm{init}$ and $\sigma_T\in X_\mathrm{goal}$.}\\ \\
Let a cost function $c(\cdot)$ computes a cost of a given path $\sigma$ in
terms of a summation of Euclidean distances between all the consecutive states in $\sigma$. Let a set of all feasible path solutions to a given planning problem be denoted as $\Pi$. The optimality problem of motion planning is then to find the optimal, feasible, path solution $\sigma^* \in \Pi$ that has a minimum cost among all other feasible path solutions i.e.,\\ \\
\textbf{Problem 2 (Optimal Path Planning)}\textit{ Assuming that multiple solutions to Problem 1 exists, find a path $\sigma^*\in \Pi$ such that $c(\sigma^*)=\{\mathrm{min}_{\sigma \in \Pi} c(\sigma)\}$.}\\\\
Let $\Omega \subset X_\mathrm{free}$ be a potential region containing optimal/near-optimal path solution. The problem of \textit{adaptive sampling}, also known as biased sampling, is to generate collision-free samples $x \in \Omega$ such that SMPs compute the optimal path $\sigma^*$ in a least possible time $t \in \mathbb{R}$. The problem of adaptive sampling is formalized as follow.\\\\
\textbf{Problem 3 (Adaptive Sampling)}\textit{ Given a planning problem $\{x_\mathrm{init},X_\mathrm{goal}, X\}$, generate samples $x\in \Omega$, where $\Omega \subset X_\mathrm{free}$, such that the sampling-based motion planning methods compute optimal path solution $\sigma^*$ in a least-possible time $t\in \mathbb{R}$.}

\section{Informed Neural Sampler}
This section presents our novel informed neural sampling algorithm called DeepSMP\footnote{Supplementary material is available at
sites.google.com/view/deepsmp}. It comprises two neural modules. The first module is an autoencoder which learns an invariant and robust feature space to embed a point cloud data from obstacle space. The second module is a stochastic DNN which takes obstacles encoding, start and goal configurations to generate samples incrementally for SMPs during online execution. Note that any SMP can utilize these informed samples for rapid convergence to the optimal solution and that the method works for unseen environments via the obstacle space encoding. The following sections describe both neural modules, online sample generation heuristic called DeepSMP, dataset collection, and hyper-parameters initialization.  
\subsection{Obstacle Encoding}
A Contractive AutoEncoder (CAE) is used to learn a latent-space embedding $Z$ of a raw point cloud data $\boldsymbol{x_\mathrm{obs}} \subset X_\mathrm{obs}$ (see Fig. 1 (a)). The encoder and decoder functions of CAE are denoted as $f(\boldsymbol{x_\mathrm{obs}}; \boldsymbol{\theta}^e)$ and $g(f(\boldsymbol{x_\mathrm{obs}}); \boldsymbol{\theta}^d)$, respectively, where $\boldsymbol{\theta}^e$ and $\boldsymbol{\theta}^d$ are parameters of their corresponding approximating functions. CAE is trained through unsupervised learning using the following objective function.
\begin{equation}
L_\mathrm{CAE}\big(\boldsymbol{\theta}^e,\boldsymbol{\theta}^d\big)= \cfrac{1}{N_\mathrm{obs}}\sum_{\boldsymbol{x}\in D_\mathrm{obs}}||\boldsymbol{x}-g(f(\boldsymbol{x}))||^2  + \lambda \sum_{ij} (\theta^e_{ij})^2
\end{equation}
where $||\boldsymbol{x}-g(f(\boldsymbol{x}))||^2$ is a reconstruction loss, and $\lambda \sum_{ij} (\theta^e_{ij})^2$ is a regularization term with a coefficient $\lambda$. Furthermore, $D_\mathrm{obs}$ contains a dataset of point clouds $\boldsymbol{x_\mathrm{obs}} \subset X_\mathrm{obs}$ from $N_\mathrm{obs} \in \mathbb{N}$ different workspaces. The regularization term allows the feature space $Z:=f(\boldsymbol{x_\mathrm{obs}})$ to be contractive in the neighborhood of the training data which results in an invariant and robust feature learning \cite{rifai2011contractive}.

\subsubsection{Model Architecture}
Since the decoding function $g(f(\boldsymbol{x_\mathrm{obs}}))$ is an inverse of encoding function $f(\boldsymbol{x_\mathrm{obs}})$, we present the architectural details of encoding unit only. 

The encoding function consists of three fully-connected linear hidden layers followed by an output linear layer. The output from each hidden layer is passed through a Parametric Rectified Linear Unit (PReLU) \cite{trottier2016parametric}.

For 2D workspaces, the input point cloud is of size $1400\times 2$ where three hidden layers transform the inputs to 512, 256 and 128 units, respectively. The output layer takes 128 units and transforms them to latent space embedding $Z$ of size 28 units. The decoding function takes the latent space embedding Z to reconstruct the raw point cloud data. 

For 3D workspaces, the hidden layers 1, 2 and 3 transform the input point cloud $1400 \times 3$ to 786, 512, and 256 hidden units, respectively. Finally, the output layer transforms the 256 units from preceding hidden layer to a latent space of size 60 units.

\subsection{Deep Sampler}
Deep Sampler is a stochastic feedforward deep neural network with parameters $\boldsymbol{\theta}$. It takes obstacles encoding $Z$, robot state $x_t$ at step $t$, and goal state $x_T$ to produce a next state $\hat{x}_{t+1} \in X_\mathrm{free}$ that would take a robot closer to the goal region (see Fig. 1(b)) i.e.,
\begin{equation}
\hat{x}_{t+1}=\mathrm{DeepSampler}((x_t,x_T,Z);\boldsymbol{\theta})
\end{equation}

We use RRT* \cite{karaman2011sampling} to produce near-optimal paths to train DeepSMP. The training paths are in the form of a tuple i.e., $\sigma^*=\{x_0,x_1,\cdots,x_T\}$, such that the path formed by connecting all following states in $\sigma^*$ is a feasible solution. The training objective is to minimize mean-squared-error (MSE) between the predicted states $\hat{x}_{t+1}$ and the actual states $x_{t+1}$ given by RRT*, i.e.,
   
\begin{equation}
L_\mathrm{MSE}(\boldsymbol{\theta})=\cfrac{1}{N_p} \sum^{\hat{N}}_j \sum^{T-1}_{i=0} ||\hat{x}_{j,i+1}-x_{j,i+1}||^2,
\end{equation}
where $N_p \in \mathbb{N}$ corresponds to the total number of paths $\hat{N}$ times their path lengths. 
  
\subsubsection{Model Architecture}
Deep Sampler is a twelve-layer deep neural network where each hidden layer is a sandwich of a linear layer, PReLU \cite{trottier2016parametric} and Dropout ($p$) \cite{srivastava2014dropout} with an exception of last hidden layer which does not contain Dropout ($p$). The twelveth layer is an output layer which takes hidden units from preceding layer and transforms them to the desired output size which is equal to the dimension of robot configurations. The configurations for the 2D point-mass robot, 3D point-mass robot, rigid-body and 6 DOF robot have dimensions 2, 3, 3 and 6 respectively. For all presented problems, except planning of 6 DOF robot, the input to Deep Sampler is given by concatenating the obstacles' representation $Z$, robot's current state $x_t$ and goal state $x_T$. For 6 DOF, we assume a single environment, therefore, the input to Deep Sampler comprises of current state $x_t$ and goal state $x_T$ only. 

\subsection{Online Execution of DeepSMP}
During the online phase, we use our trained obstacle encoder and DeepSampler to generate random samples for a given SMP. Fig. 1 shows the flow of information between encoder $f(\boldsymbol{x_\mathrm{obs}})$ and DeepSampler. Algorithm 1 outlines DeepSMP which combines our informed neural sampler with any classical SMP such RRT*.

\begin{algorithm}[t]
\DontPrintSemicolon 
\textbf{Initialize} SMP($x_\mathrm{init},x_\mathrm{goal}, X$)\;
$x_\mathrm{rand} \gets x_\mathrm{init}$\;
$Z \gets f(\boldsymbol{x_\mathrm{obs}})$\;
\For{$i \gets 0$ \textbf{to} $n$} {

\If{$i <n^\mathrm{limit}$}
   {
   $x_\mathrm{rand} \gets \mathrm{DeepSampler}\big(Z, x_\mathrm{rand}, x_\mathrm{goal}\big)$\;
   	  
   }
 \Else
 {
    $x_\mathrm{rand} \gets \mathrm{RandomSampler}()$\;

 }

    $\sigma \gets \mathrm{SMP}\big(x_\mathrm{rand}\big)$\;
	
   \If{$x_\mathrm{rand} \in X_\mathrm{goal}$}
   {
   $x_\mathrm{rand} \gets x_\mathrm{init}$\;
   	  
   }
   }

   \If{$\sigma_T \in X_\mathrm{goal}$}
   {
   \Return{$\sigma$}\;
   	  
   }

\Return{$\varnothing$}\;
\caption{DeepSMP($x_\mathrm{init}, x_\mathrm{goal}, \boldsymbol{x_\mathrm{obs}} $)}
\label{algo:}
\end{algorithm}

Algorithm 1 starts by initializing a given SMP (Line 1). The obstacles encoder $f(\boldsymbol{x_\mathrm{obs}})$ provides an encoding $Z$ of a raw point cloud data from $X_\mathrm{obs}$ (Line 3). DeepSMP algorithm runs for $n \in \mathbb{N}$ iterations (Line 4). DeepSampler incrementally generates samples between given start and goal configurations until $i <n^\mathrm{limit}$ (Line 5-6), where $n^\mathrm{limit} < n$. Upon reaching a given goal configuration, DeepSampler is executed again to produce samples from a given start configuration to the goal configuration by re-initializing random sample $x_\mathrm{rand}$ to $x_\mathrm{init}$ (Lines 10-11). After $n^\mathrm{limit}$ iterations, DeepSMP switches to random sampling (Line 7-8) to ensure completeness guarantees of an underlying SMP. Note that $\sigma$ is a feasible path solution returned by SMP. The path $\sigma$ is continually optimized for a given cost function $c(\cdot)$ for a given number of iteration $n$. Finally, after $n$ iterations, a feasible, optimized path solution $\sigma$, if one exists, is returned as a solution to a given planning problem (Lines 12-13).

\subsection{Data Collection}
The data collection consists of creating a random set of workspaces, sampling collision-free start and goal configurations in those workspaces, and generating paths using a classical motion planner for every start and goal pair. The following sections describe the procedure to create workspaces, start and goal pairs, and near-optimal paths.
\subsubsection{Workspaces}
Many different 2D and 3D workspaces were generated by randomly placing various quadrilateral blocks without repetition in the operating region of $40 \times 40$ and $40 \times 40 \times 40$, respectively. Each random placement of the obstacle blocks led to a different workspace. 
\subsubsection{Start and goal configuration}
For each generated workspace, a number of start and goal configurations were sampled randomly from its obstacle-free space.
\subsubsection{Near-optimal paths}
Finally, for each generated start and goal pair within all workspaces, a feasible, near-optimal path was generated using the RRT* motion planner. 

Complete dataset comprised 110 different workspaces for the presented scenarios in the results section i.e., simple 2D (s2D), complex 2D (c2D), complex 3D (c3D), and rigid-body (rigid). The training dataset contained 100 workspaces with 4000 training paths in every workspace. There were two types of test datasets. The first test dataset comprised already seen 100 workspaces with 200 unseen start and goal configurations in each of the workspaces. The second test dataset comprised entirely unseen 10 workspaces where each contained 2000 unseen start and goal configurations. For rigid-body case, the range of angular configuration was scaled to the range of positional configurations, i.e., $-20$ to $20$, for training and testing. In case of 6 DOF robot, we consider only a single environment thus no environment encoding is included, and only start and goal configurations are sampled to collect example trajectories (50,000) from collision-free space to train our feedforward neural network (DeepSampler). The test scenario for 6 DOF robot is to generate paths for unseen start and goal pairs.

\subsection{Hyper-parameters}
DeepSMP neural models were trained in mini-batches using Adagrad \cite{duchi2011adaptive} optimizer with a learning rate of $0.1$. CAE was trained on raw point cloud data from $N_\mathrm{obs}=30,000$ different workspaces which were generated randomly as described earlier. The regularization coefficient $\lambda$ was set to $10^{-3}$. For DeepSampler, Dropout probability $p$ was kept constant to 0.5 for both training and testing. The number $n^\mathrm{limit}$ is set to the number of nodes in the longest path available in the training data. For RRT*, gamma of ball-radius was set to 1.6 whereas tree extension step sizes for point-mass and rigid-body were kept at 0.01 and 0.9, respectively. Finally for the 6-DOF robot, we use OMPL's RRT* and ROS with their default parameter settings for path generation.  
\begin{figure}[t]
    \centering
    \begin{subfigure}[b]{0.23\textwidth}
      \fbox{\includegraphics[height=3.9cm,width=3.9cm]{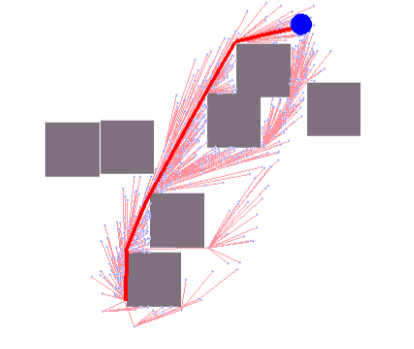}}
       \caption{$n=523, t=0.96s$ }
    \end{subfigure}
    \begin{subfigure}[b]{0.23\textwidth}
       \fbox{\includegraphics[height=3.9cm,width=3.9cm]{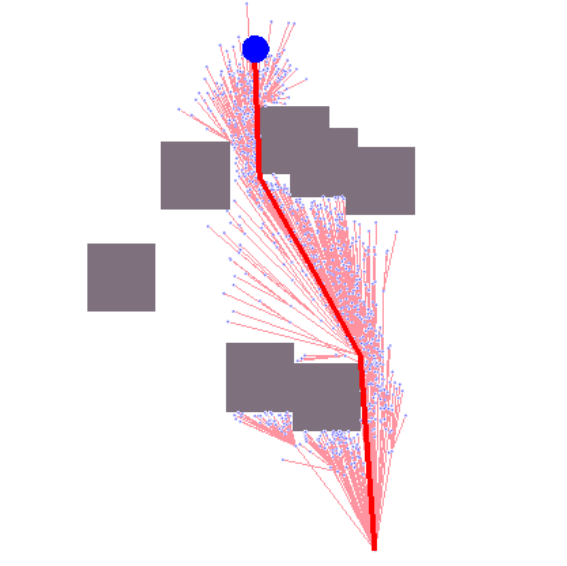}}
        \caption{$n=418, t=0.82s$}
    \end{subfigure}
    \caption{DeepSMP in simple 2D environments (s2D).}\label{s2D}
\end{figure}
\begin{figure*}[t]
    \centering
    \begin{subfigure}[b]{0.24\textwidth}
      \fbox{\includegraphics[height=3.85cm,width=4.1cm]{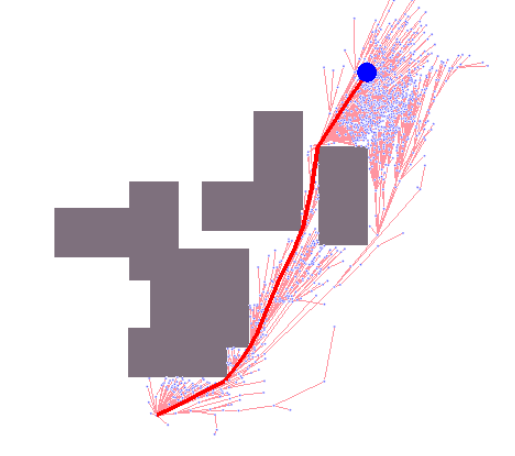}}
       \caption{$n=821, t=1.21s$}
    \end{subfigure}
     \begin{subfigure}[b]{0.24\textwidth}
      \fbox{\includegraphics[height=3.85cm,width=4.1cm]{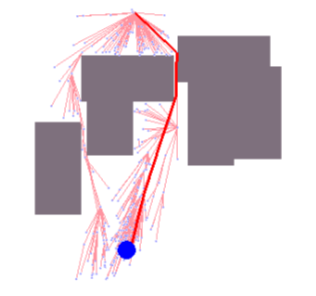}}
        \caption{$n=913, t=0=1.32s$}
    \end{subfigure}
    \begin{subfigure}[b]{0.24\textwidth}
       \fbox{\includegraphics[height=3.85cm,width=4.1cm]{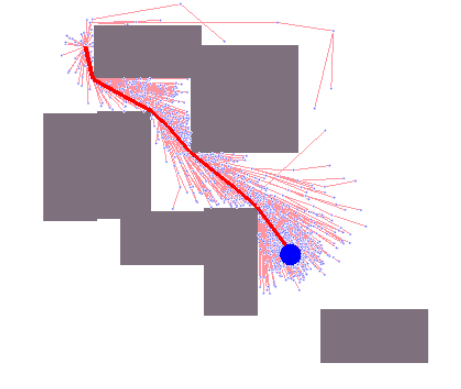}}
        \caption{$n=1087, t=1.48s$}
    \end{subfigure}
    \begin{subfigure}[b]{0.24\textwidth}
       \fbox{\includegraphics[height=3.85cm,width=4.1cm]{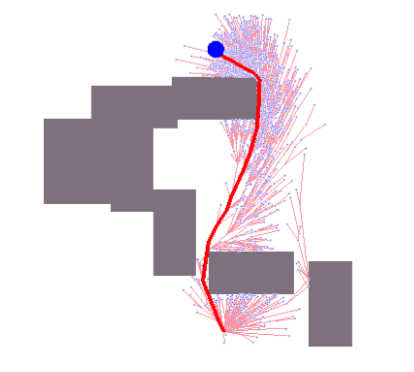}}
        \caption{$n=785, t=1.02s$}
    \end{subfigure} 
    \caption{ DeepSMP in complex 2D environments (c2D). The path and goal are indicated in red and blue colors, respectively.}\label{c2D}
\end{figure*}
\section{Results}
This section presents the results of DeepSMP for the motion planning of a point-mass robot, rigid-body, and  Universal 6 DOF robot (UR6) in both 2D and 3D environments. All experiments were carried out on a computer with 3.40GHz$\times$ 8 Intel Core i7 processor with a 16 GB RAM and GeForce GTX 1080 GPU. DeepSMP, implemented in PyTorch, was compared against Informed-RRT* and BIT* implemented in Python. In the following results, the datasets $\mathrm{seen}$-$X_\mathrm{obs}$ and $\mathrm{unseen}$-$X_\mathrm{obs}$ comprised 100 workspaces seen by DeepSMP during training and 10 workspaces not seen by DeepSMP during training, respectively. Both test datasets $\mathrm{seen}$-$X_\mathrm{obs}$ and $\mathrm{unseen}$-$X_\mathrm{obs}$ contained 200 and 2000 unseen start and goal configurations, respectively, for every workspace. Note that every combination of either seen or unseen environment with unseen start and goal pair constitutes a new planning problem, i.e., not seen by DeepSMP during training. For each planning problem, we ran 20 trials of all presented SMPs to calculate the mean computational time.
\begin{figure}[t]
    \centering
    \begin{subfigure}[b]{0.23\textwidth}
      \includegraphics[width=4.3cm]{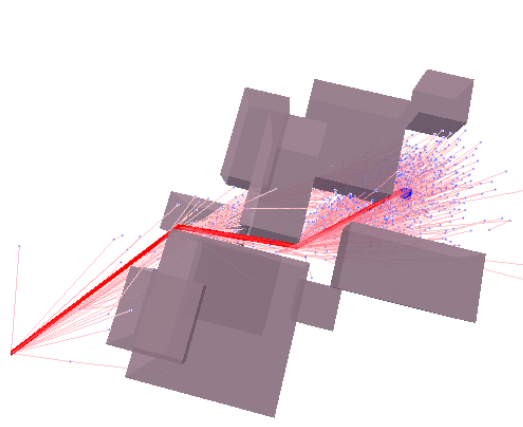}
       \caption{$n=521, t=0.79s$}
    \end{subfigure}
    \begin{subfigure}[b]{0.23\textwidth}
       \includegraphics[width=4.3cm]{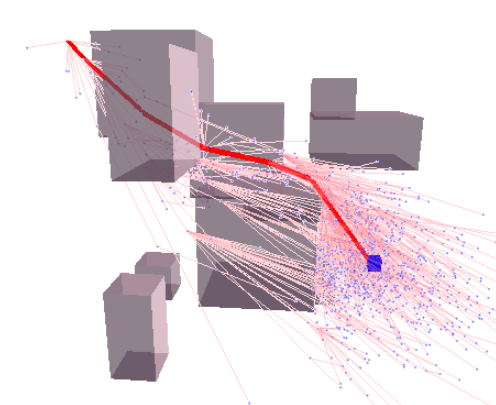}
        \caption{$n=974, t=1.32s$}
    \end{subfigure}
    \caption{DeepSMP generating samples in complex 3D environments (c3D). The obstacles, indicated as blocks in beige color, are made slightly transparent to display path profiles behind them.}\label{c3D}
\end{figure}
Figs. \ref{s2D}-\ref{r2D} show different example scenarios named as simple 2D (s2D), complex 2D (c2D), complex 3D (c3D) and rigid-body (rigid) where DeepSMP with underling RRT* method is planning motions. The mean computational time (in seconds) and iterations took by DeepSMP for each scenario are denoted as $t$ and $n$, respectively. 

Table I presents the mean computational time comparison of DeepSMP with an underlying RRT* SMP against Informed-RRT* and BIT* for computing near-optimal paths in different environments s2D, c2D, c3D and rigid. Note that, unbiased RRT* method is not included in the comparison as the computation time of RRT*, for computing near-optimal paths, is much higher than all presented algorithms. We report the mean ($t_\mathrm{mean}$), maximum ($t_\mathrm{max}$), and minimum ($t_\mathrm{min}$) time taken by an algorithm in every environment. It can be seen that in all test cases, the mean computation time of DeepSMP:RRT* remained consistently around 2 seconds. However, the mean computation time of Informed-RRT* and BIT* increases significantly as the dimensionality of the planning problem increases slightly. Furthermore, the rightmost column presents the ratio of mean computational time of BIT* to DeepSMP, and it is observed that on average, our method is at least 7 times faster than BIT*, the current state-of-art motion planner. 

From experiments presented so far, it is evident that BIT* outperforms Informed-RRT*, therefore, in the following experiments only DeepSMP and BIT* are compared. Fig. \ref{tc} compares the mean computation time of DeepSMP: RRT* and BIT* in two test cases, i.e., $\mathrm{seen}$-$X_\mathrm{obs}$ and $\mathrm{unseen}$-$X_\mathrm{obs}$. It can be observed that the mean computation time of DeepSMP stays around 2 seconds irrespective of the given problem's dimensionality. Furthermore, the mean computational time of BIT* not only fluctuates but also increases significantly as the dimensionality of the planning problem increases slightly.
\begin{figure}
    \centering
      \fbox{\includegraphics[height=5.0cm]{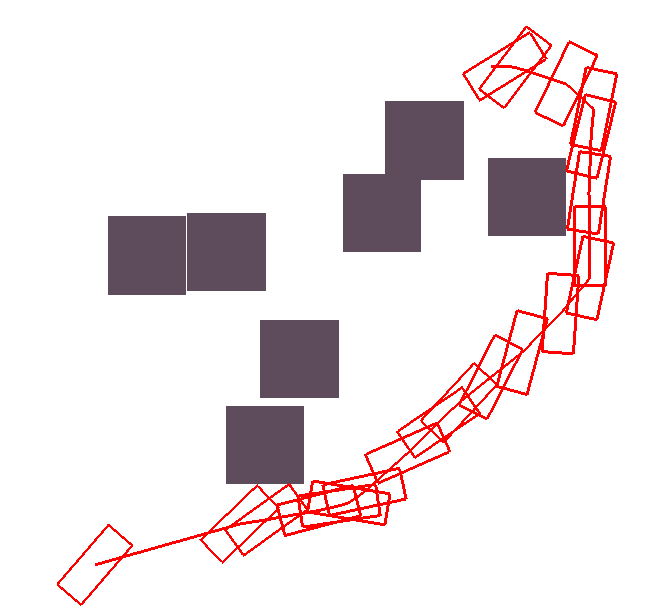}}
    \caption{DeepSMP computed optimal path for a rigid body in 1.12 seconds.}\label{r2D}
\end{figure}
Finally, Fig. \ref{ur6} shows DeepSMP planning motions for a Universal 6-DOF robot.  In Fig. \ref{ur6} (a), the robotic manipulator is at the start configuration whereas its target configuration is symbolized as a shadowed region. Fig. \ref{ur6} (b) shows the traces of a path planned by DeepSMP for the given start and goal pair. In this problem, the mean computational times taken by DeepSMP and BIT* are 1.7 and 48.8 seconds, respectively, which makes DeepSMP around 28 times faster than BIT*.
\begin{table*}
\centering 
\begin{tabular}{|c|c|c|c|c|c|c|c|c|c|c|c|}\hline
\multirow{2}{*}{Environment}& \multirow{2}{*}{Test case}&\multicolumn{3}{c|}{DeepSMP:RRT* }&\multicolumn{3}{c|}{Informed-RRT*}&\multicolumn{3}{c|}{BIT*}& \multirow{2}{*}{\scriptsize{$\cfrac{\mathrm{BIT}: t_\mathrm{mean}}{\mathrm{DeepSMP}:t_\mathrm{mean}}$}}  \\\cline{3-11}
&& $t_\mathrm{mean}$& $t_\mathrm{max}$ & $t_\mathrm{min}$& $t_\mathrm{mean}$ &$t_\mathrm{max}$& $t_\mathrm{min}$& $t_\mathrm{mean}$ &$t_\mathrm{max}$& $t_\mathrm{min}$&  \\ \hline \hline

\multirow{2}{*}{Simple 2D (s2D)}& Seen $X_\mathrm{obs}$ & 0.90& 1.09 & 0.78& 9.61 &11.90& 3.21& 4.62 &10.68& 1.79& 5.13 \\ \cline{2-12} 

& Unseen $X_\mathrm{obs}$ & 0.92& 1.00 & 0.87& 9.89 &9.24& 6.19& 5.05 &3.68& 1.45&5.49  \\ \hline

\multirow{2}{*}{Complex 2D (c2D)}& Seen $X_\mathrm{obs}$ & 1.62& 2.19 & 1.09& 10.81 &14.30& 7.11&  6.56 &12.02& 3.22&4.05   \\ \cline{2-12} 

& Unseen $X_\mathrm{obs}$ & 1.46& 2.11 & 1.00& 11.21 &12.51& 4.36& 5.72 &9.89& 2.92&3.92  \\ \hline

\multirow{2}{*}{Complex 3D (c3D)}& Seen $X_\mathrm{obs}$ & 1.16& 1.72 & 0.72 & 18.15 &74.50& 16.69& 16.92 &49.03& 6.19&14.68  \\ \cline{2-12} 

& Unseen $X_\mathrm{obs}$ & 1.36& 1.96 & 0.94& 18.43 &49.37& 12.17& 15.96 &26.16& 12.53&11.74  \\ \hline

\multirow{2}{*}{Rigid-body (rigid)}& Seen $X_\mathrm{obs}$ & 1.61& 2.65 & 0.71& 42.78 &209.12& 38.34& 16.01 &34.64& 7.81&9.94  \\ \cline{2-12} 

& Unseen $X_\mathrm{obs}$ & 1.72& 2.81 & 1.01 & 43.74 &188.63& 27.45 & 16.61 &34.65& 7.85&9.65  \\ \hline
\end{tabular}
\caption{Time comparison (in seconds) of DeepSMP against Informed-RRT* and BIT* on two test datasets.}
\end{table*}

\section{Discussion}
\subsection{Stochasticity through Dropout}
Our stochastic feedforward DeepSampler uses Dropout \cite{srivastava2014dropout} in every layer except the last two layers during both offline and online execution. Dropout is applied layer-wise to a neural network, and it drops each unit in the hidden layer with a probability $p \in [0,1]$. In our models, the dropped out units are indicated as dotted circles in Fig. 1. Thus, the resulting neural network is a sliced version of the original deep model, and in every iteration during online execution, a different model emerges through randomly dropping some hidden units. These perturbations in DeepSampler through Dropout enables DeepSMP to generate different samples in the region likely to contain path solutions.
\begin{figure}[h!]
    \centering
    \begin{subfigure}[b]{0.55\textwidth}
       \includegraphics[height=6.0cm, width=9.2cm]{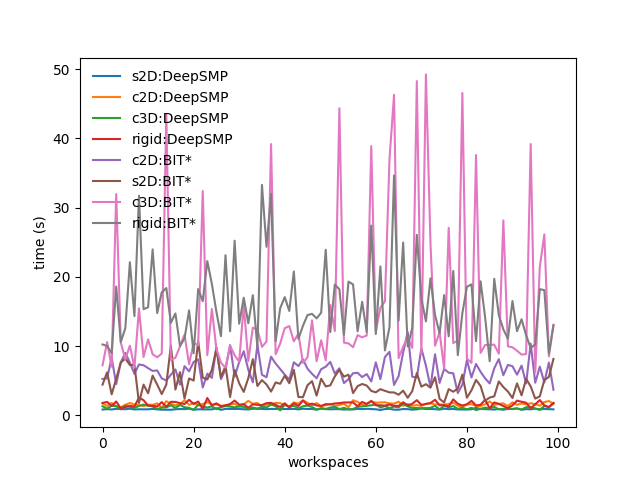}
        \caption{Test-case 1: seen-$X_\mathrm{obs}$}
    \end{subfigure}
        \begin{subfigure}[b]{0.55\textwidth}
       \includegraphics[height=6.0cm, width=9.2cm]{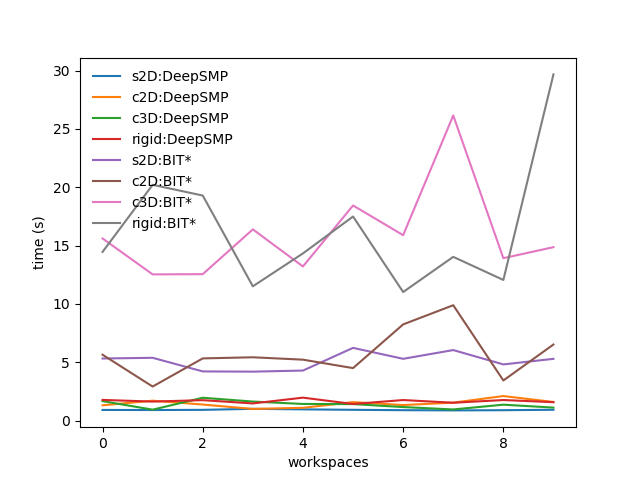}
    \caption{Test-case 2: unseen-$X_\mathrm{obs}$}
    \end{subfigure}
    \caption{Computational time comparison of DeepSMP:RRT* and BIT* on test datasets. The plots show DeepSMP is more consistent and faster than BIT* in all test cases. }\label{tc}
\end{figure} 
\subsection{Bidirectional Sampling}
Since our method incrementally generates samples, it can be easily extended to produce samples for bidirectional SMPs such as IB-RRT* \cite{qureshi2015intelligent}. To do so, treat both start and goal configuration as random variables $x_\mathrm{rand1}$ and $x_\mathrm{rand2}$, respectively, and swap their roles by the end of every iteration in Algorithm 1. This way, two trees in bidirectional SMPs can be made to march towards each other to rapidly compute end-to-end collision-free paths.
\subsection{Completeness}
SMPs ensure \textit{probabilistic completeness}. Let $V^\mathrm{SMP}_n$ denotes the tree vertices of SMP after $n \in \mathbb{N}$ iterations. Since all SMPs begin to build a tree from initial robot state $x_\mathrm{init}$ i.e., $V^\mathrm{SMP}_0=x_\mathrm{init}$, and randomly explore the entire configuration space by forming a connected tree as $n$ approaches to infinity, they guarantee  probabilistic completeness i.e.,
\begin{equation}
\mathrm{lim}_{n \rightarrow \infty} \mathbb{P}(V_n^{\mathrm{SMP}} \cap X_\mathrm{goal}\neq \varnothing)=1
\end{equation}

DeepSMP also starts generating a connected tree from $x_\mathrm{init}$ and after exploring a region that most likely contains a path solution for $n^\mathrm{limit}$ iteration, it switches to uniform random sampling (see Algorithm 1). Therefore, if $n^\mathrm{limit} \ll n$, DeepSMP also ensures \textit{probabilistic completeness} i.e., as the number of iterations $n$ approach to infinity, the probability of DeepSMP finding a path solution, if one exists, approaches to one:  
\begin{equation}
\mathrm{lim}_{n \rightarrow \infty} \mathbb{P}(V_n^{\mathrm{DeepSMP}} \cap X_\mathrm{goal}\neq \varnothing)=1
\end{equation}  
\subsection{Asymptotic Optimality}
RRT* and its variants are known to ensure asymptotic optimality i.e., as the number of iterations $n$ approaches to infinity/large-number, the probability of finding a minimum cost path solution reaches to one. This property comes from incrementally rewiring the RRT graph connections such that the shortest path is asymptotically guaranteed in RRT*. It is proposed that if the underlying SMP of DeepSMP is RRT* or any optimal variant of RRTs, DeepSMP is guaranteed to be asymptotic optimal. This follows from the fact that DeepSMP samples a selective region for fixed number of iterations and switches to uniform randoms sampling afterwards. Thus if the number of iterations goes infinity, through incremental rewiring of DeepSMP graph, the asymptotic optimality is also guaranteed.  
\begin{figure}[t]
    \centering
    \begin{subfigure}[b]{0.23\textwidth}
      \includegraphics[height=4.2cm]{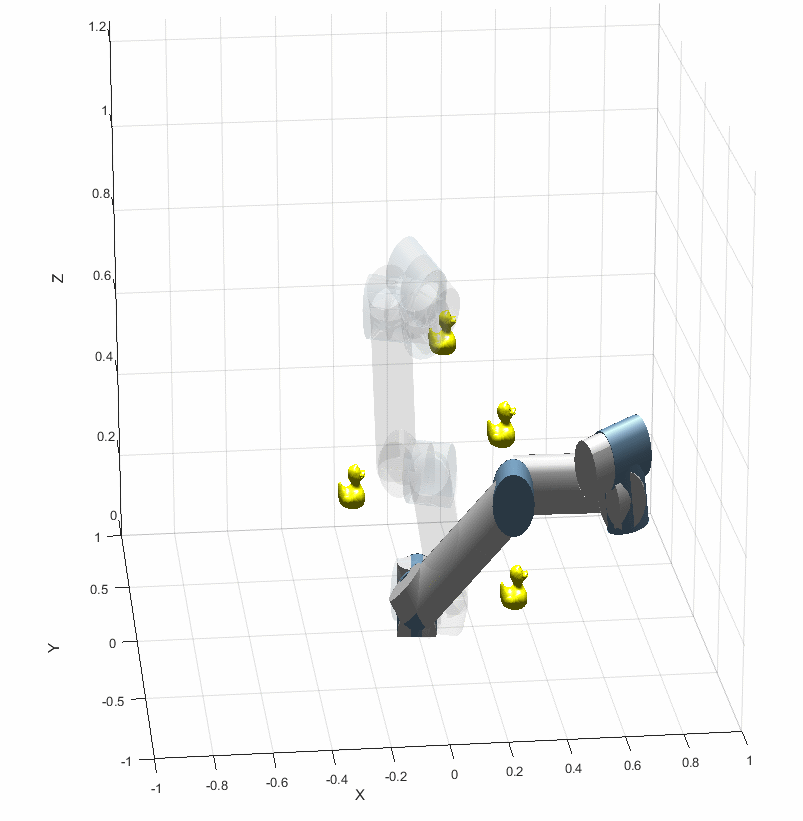}
       \caption{}
    \end{subfigure}
    \begin{subfigure}[b]{0.23\textwidth}
       \includegraphics[height=4.2cm]{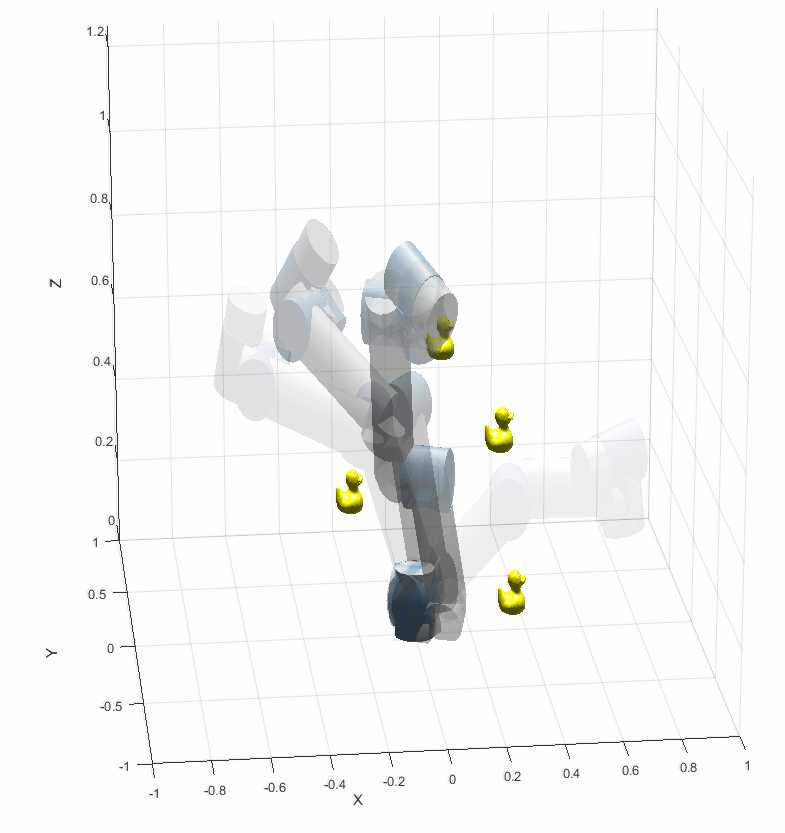}
        \caption{}
    \end{subfigure}
    \caption{DeepSMP with RRT* planning motions for a 6 DOF manipulator. Fig (a) indicates the robot at start configuration and the goal configuration is indicated as a shadowed region. Fig (b) shows the path traces followed by the robot. In this problem, the mean computational times of DeepSMP and BIT* are 1.7 and 48.8 seconds, respectively, which makes DeepSMP about 28 times faster than BIT*.}\label{ur6}
\end{figure} 
\subsection{Computational Complexity}
A forward pass through a deep neural network is known to exhibit $O(1)$ complexity. It can be seen in Algorithm 1 that adaptive samples are generated incrementally by forward passing through our stochastic DeepSampler. Hence, the proposed neural sampling method does not add any extra computational overhead to any underlying SMP for path generation. Thus, the computational complexity of DeepSMP method will essentially be the same as underlying SMP in Algorithm 1. For instance, as in our case, RRT* is an underlying SMP method, therefore, in presented experiments, the computational complexity of DeepSMP is $O(nlogn)$, where $n$ is the number of nodes in the tree.

\section{Conclusions and Future work }
In this paper, we present a deep neural network based sampling method called DeepSMP which generates samples for Sampling-based Motion Planning algorithms to compute optimal paths rapidly and efficiently. The proposed method 1) adaptively samples a selective region of a configuration space that most likely contains an optimal path solution, 2) combined with SMP methods consistently demonstrate mean execution time of about 2 second in all presented experiments, and 3) generalizes to new unseen environments.

In our future work, we plan to propose an incremental online learning method that begins with an SMP method, and trains DeepSMP simultaneously to gradually switch from uniform sampling to adaptive sampling. To speed up the incremental online learning process, we plan to propose a method that prioritizes experiences to learn from selectively fewer training examples. 


\bibliographystyle{IEEEtran}
\bibliography{references}
\nocite{*}
\end{document}